\documentclass[10pts, twocolumn]{article}
\usepackage{graphicx}
\usepackage[numbers]{natbib}
\usepackage{amsmath}
\usepackage{amsfonts}
\usepackage{cite}
\usepackage{hyperref}
\usepackage[margin=1.1in]{geometry}
\usepackage{url}

\begin{document}

\title{\textbf{Background Hardly Matters: Understanding Personality Attribution in Deep Residual Networks}}

\author{Gabri\"{e}lle Ras$^1$, Ron Dotsch$^2$, Luca Ambrogioni$^1$, Umut G\"{u}\c{c}l\"{u}$^1$, Marcel A. J. van Gerven$^1$ 
\\ \\ $^1$ Radboud University, Donders Institute for Brain, Cognition and Behaviour, \\ Nijmegen, the Netherlands \\ \texttt{\{g.ras, l.ambrogioni, u.guclu, m.vangerven\}@donders.ru.nl} \\ \\ $^2$ Anchormen, Amsterdam, the Netherlands \\
\texttt{r.dotsch@anchormen.nl} }
\date{}

\maketitle
\begin{abstract}
Perceived personality traits attributed to an individual do not have to correspond to their actual personality traits and may be determined in part by the context in which one encounters a person. These apparent traits determine, to a large extent, how other people will behave towards them. Deep neural networks are increasingly being used to perform automated personality attribution (e.g., job interviews). It is important that we understand the driving factors behind the predictions, in humans and in deep neural networks. This paper explicitly studies the effect of the image background on apparent personality prediction while addressing two important confounds present in existing literature; overlapping data splits and including facial information in the background. Surprisingly, we found no evidence that background information improves model predictions for apparent personality traits. In fact, when background is explicitly added to the input, a decrease in performance was measured across all models.
\end{abstract}

\section{Introduction}
In personality research, the Big Five model (B5)~\citep{goldberg1993structure} is the dominant paradigm used to measure various aspects of personality and how these aspects relate to an individual's happiness, choices and social behavior~\citep{funder2001personality}. The B5 is a taxonomy for personality traits, based on common language descriptors and consists of the five personality traits \textit{openness, conscientiousness, extraversion, agreeableness} and \textit{neuroticism}, represented by the acronym \textit{OCEAN}. The B5 has been shown to be a reliably stable model across a wide range of situations and cultures~\citep{deary2009trait}. Oftentimes in real life, a person does not have direct access to the true composition of another person's B5 traits. Instead, indirect cues are used to attribute apparent traits to that other person~\citep{brunswik1947systematic}. Apparent personality traits attributed to an individual do not have to correspond to their actual personality traits, however, these apparent traits determine, to a large extent, how other people will behave towards them~\citep{uleman2008spontaneous}. It has been shown that the face is an important information source when humans make (potentially actionable) personality and character judgements about others, e.g., when forming first impressions~\citep{willis2006first} or deciding which candidate to vote for~\citep{olivola2010elected}. In general there is a large number of papers showing \textit{``both correlational and causal evidence linking facial appearance to a variety of important social outcomes"}~\citep{todorov2015social}. Of course, the face is not the only source of information and there is evidence that the voice and general appearance, e.g., body posture, also influence personality judgements~\citep{olivola2010elected, naumann2009personality, ekman1980relative}. There is some evidence that humans can use objects/items that belong to the other person to base trait attribution on, such as clothing~\citep{howlett2013influence} and even bedroom objects~\citep{poggio2013inferences}. Studies in market research often find that brands are associated with certain personality traits~\citep{aaker1995brand}. In the related field of emotion perception, the role of environmental cues has been studied more extensively and they show that the background scene influences emotional judgements in humans~\citep{barrett2010context, righart2008recognition, barrett2011context, aviezer2008angry}. Yet, it remains unclear to what extent physical object and environmental cues can communicate a person's personality traits and how reliable these cues are when attributing apparent traits to other people.

\subsection{Automated Personality Attribution}
The field of personality computing studies and develops computational methods that, among other things, attribute \textit{apparent} personality traits to individuals~\citep{vinciarelli2014survey}. By learning from a large dataset of human annotated data, these methods can, by proxy, serve as models of human behavior. A popular method is the Deep Neural Network (DNN)\footnote{In this paper we use the terms "DNN" and "model" interchangeably}. A recent survey on personality attribution using DNNs indicate that researchers tend to feed the DNNs predominantly with data that contains a face~\citep{junior2018first}. The following factors might contribute to this trend; 1) There is an abundance of evidence supporting the role of facial cues in personality attribution and insufficient evidence that environmental cues in the background image influences attribution. 2) Using only the face vs. other regions in the image serves as a form of dimensionality reduction of the input data, giving rise to additional benefits such reduced training time and computational resources. 3) In the earlier days of social signal processing, the background of the image was being discarded because it was considered a source of noise for the algorithms being used~\citep{poria2017review}. 
\\
Efforts based on extracting information from additional cues often take the approach of feeding different modalities (audio, text) to the DNN and we generally find that the DNN makes better predictions when it receives information from multiple sources~\citep{junior2018first}. However, there is a lack of research providing insight into how integrating visual environmental cues from the background of the image can aid the attribution of personality traits. Given the recent advances in DNNs and their ability to automatically extract relevant features from the input, we might want to reconsider leaving out the background.

\subsection{Confounds in First Impressions v2 Dataset}
\label{sec:chalearn}
Fairly recently, the ChaLearn First Impressions v2 dataset has been publicly released~\citep{ponce2016chalearn}. This dataset is a large collection of YouTube vlog clips\footnote{A vlog is a video blog}, annotated by crowdsourcing with Amazon Mechanical Turk\footnote{\url{https://www.mturk.com/}}. Each video depicts one person speaking directly to the camera, often in a home environment, making it an ideal dataset with which to investigate the influence of environmental cues on trait attribution. About $75\%$ of the original full length videos are split into two to six 15 second clips. More dataset details are provided in Section \ref{sec:dataset}. Currently this is the largest video personality dataset in existence. The dataset was released in the context of a conference competition and has resulted in a number of papers in which the role of the environmental cues was investigated~\citep{escalante2018explaining, guccluturk2016deep, gucluturk2017visualizing, wei2018deep, gurpinar2016multimodal, gurpinar2016combining}. However, we discovered two confounds that severely limit the conclusions of these studies. 
\begin{enumerate}
    \item In machine learning, it is good practice to judge the performance of a model on a set of data that the model has never seen before. This is achieved by partitioning the entire dataset into a train and test split. After the DNN has been trained, its performance is measured on this unseen portion of data, very similar to an exam at the end of a school course. However, in the First Impressions dataset this independent assessment is counfounded by the fact that $83\%$ of the clips in the test split originate from the same video as $46\%$ of clips in the train split, see Figure \ref{fig:data_overlap}. The independence of this assessment is very important because it gives us a measure of how well the DNN can generalize what it has learned and not just what it has memorized.
    \item The other confound is present in the analysis provided in the papers. In order to study the effect of environmental cues, there needs to be a condition in which only the environmental cues are provided during the learning (training) process and evaluation (testing) process of the DNN. However, not a single analysis meets this condition; in all training and testing procedures facial features were present.
\end{enumerate}

\subsection{Our Contribution}
This paper contributes to personality computing by explicitly studying the effect of visual cues in the image background on personality trait attribution, using DNNs, while addressing the existing confounds. In this paper we ask the question \textit{do environmental cues encoded in the image background significantly inform apparent personality trait attribution}? We expect to find that when background information is added to the DNN, the model performance increases. The answer to this question is very relevant; if it turns out that background information improves personality attribution then researchers in the personality computing field might want to consider methods that make use of the background information. By studying the behavior of the DNN we might also gain insight into how humans make personality attributions. 

\section{Related Work}
All of the following papers use the First Impressions dataset and the original dataset splits where $83\%$ of the clips in the test split originate from the same video as $46\%$ of clips in the train split. 

G{\"u}rp{\i}nar et al. ~\citep{gurpinar2016combining, gurpinar2016multimodal} use a combination of DNNs to extract audio, scene and facial features from the data. These features are fused together to predict the final attribution scores. The scene component of the DNN has been trained on the ImageNet dataset~\citep{ILSVRC15} which includes facial information. The ablation studies indicate that their method benefits from the additional scene features. However, scene information is always entangled with facial information. Wei et al.~\citep{wei2018deep} train various DNNs by using the information encoded in the entire frame. A visual feature importance analysis was performed on one of the higher (decision) layers of the networks and  the results suggest that DNNs pay attention to the background of the image when making predictions. However, this analysis was limited to only 12 random frames in the test split. Given that there is a large overlap between the train and test data, the results of this analysis can be a consequence of the DNNs simply recalling what they have seen during the training phase. 
G{\"u}{\c{c}}l{\"u}t{\"u}rk et al.~\citep{guccluturk2018multimodal} take into account scene information by feeding their DNNs random frame crops during training. G{\"u}{\c{c}}l{\"u}t{\"u}rk et al.~\citep{gucluturk2017visualizing} use occlusion analysis to visualize regions in the image input that are important for the predictions. The results show that the background provides important information. However, similar to the limitations of~\citep{wei2018deep}, the DNN can simply be recalling information from an image that it has seen before.

\section{Experimental setup}
\begin{figure}
    \centering
    \includegraphics[width=1.0\linewidth]{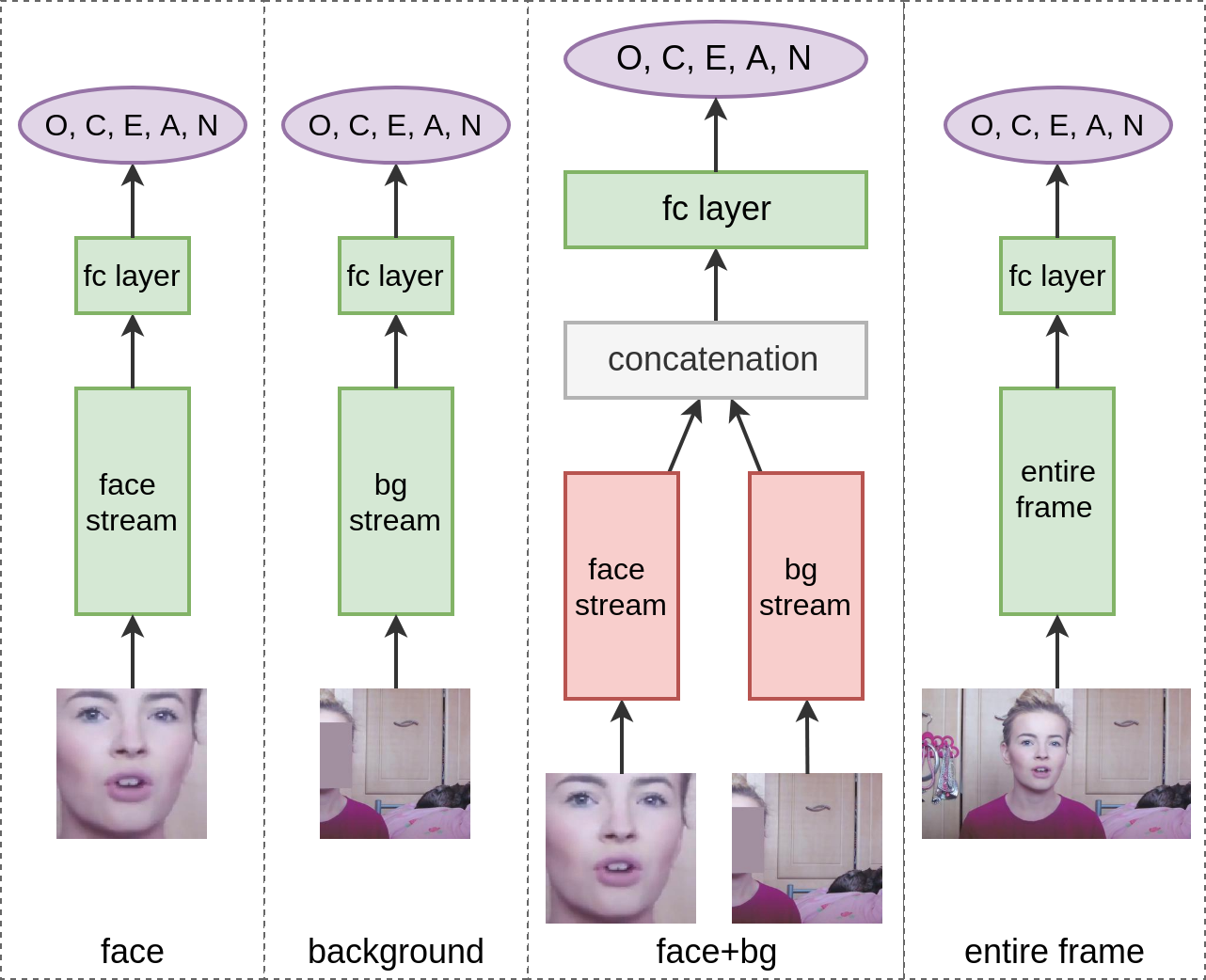}
    \caption{The general setup of the experiments. Experimental conditions \textit{face}, \textit{background} and \textit{entire frame} all use the same network architecture. In condition \textit{face+bg} the data is passed through a two-stream network, of which only the final fully connected layer is trained. Each stream adopts and freezes the weights of a previously trained face or background model.}
    \label{fig:data_models}
\end{figure}

To account for the previously mentioned confounds, we 1) create new data splits, where the clips in one split are completely independent from clips in another split, and 2) implement an experimental design enabling the study of the effect of environmental cues on the DNN attribution capabilities, see Figure \ref{fig:data_models}. All following experiments are performed using only the individual frames of the videoclips. Hence the word \textit{frame} will be used instead of videoclip. Given an input, the DNN has to predict on a continuous scale between 0 and 1 the intensity of each B5 trait. The experiments are repeated on three different DNNs. Finally the DNN predictions are compared to the actual labels and the attribution performance is measured. We hypothesize that if the background contains useful cues, these will be extracted and utilized by the DNN. As a result there should be an increase in attribution performance in conditions where the background is included in the input. 

\subsection{Dataset Details}
\label{sec:dataset}

\begin{figure}
    \centering
    \includegraphics[width=0.7\linewidth]{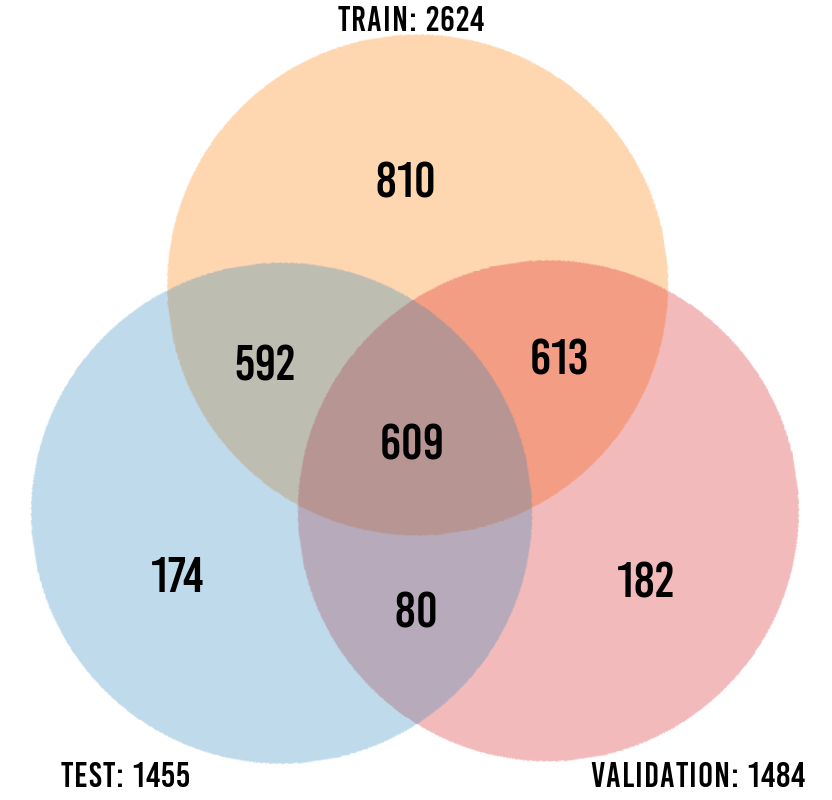}
    \caption{A Venn diagram visualizing the portion of overlapping data in the original ChaLearn First Impressions split.}
    \label{fig:data_overlap}
\end{figure}

\begin{table}
\renewcommand{\arraystretch}{1.3}
\caption{The number of videos in each data split and the number of videos per UID. All of the datasplits are independent of each other, meaning that there are no overlapping UIDs.}
\centering
\begin{tabular}{|c|c|c|c|}
\hline
\textbf{} & \textbf{\begin{tabular}[c]{@{}c@{}}number\\ of \\ videos\end{tabular}} & \textbf{\begin{tabular}[c]{@{}c@{}}unique\\ ID's \\(UID)\end{tabular}} & \textbf{vid/UID} \\ \hline
\textbf{training} & 6744 & 2060 & 3.27 \\ \hline
\textbf{testing} & 1676 & 500 & 3.35 \\ \hline
\textbf{validation} & 1580 & 500 & 3.16 \\ \hline
\end{tabular}
\label{tab:data_information}
\end{table}

The ChaLearn First Impressions v2 dataset~\citep{ponce2016chalearn} is used in all of the experiments. This dataset is a collection of 10000 HD 720p YouTube videoclips, gathered from 3060 unique videos. The duration of each videoclip is 15 seconds and has an average framerate of 30 FPS. The videoclips were annotated by Amazon Mechanical Turk workers. The labels in the dataset indicate the intensity of the perceived Big Five traits on a continuous scale between 0 and 1. Unfortunately, the original dataset splits include a rather large number overlapping dependent videoclips, see Figure \ref{fig:data_overlap}. We created new splits where the clips in one split are completely independent from clips in another split. The dataset splits can be obtained by sorting the unique video names alphabetically and then selecting sequentially the respective number of videos in each data split indicated in Table \ref{tab:data_information}.

\subsection{Experimental Conditions}
\begin{figure}
    \centering
    \includegraphics[width=1.0\linewidth]{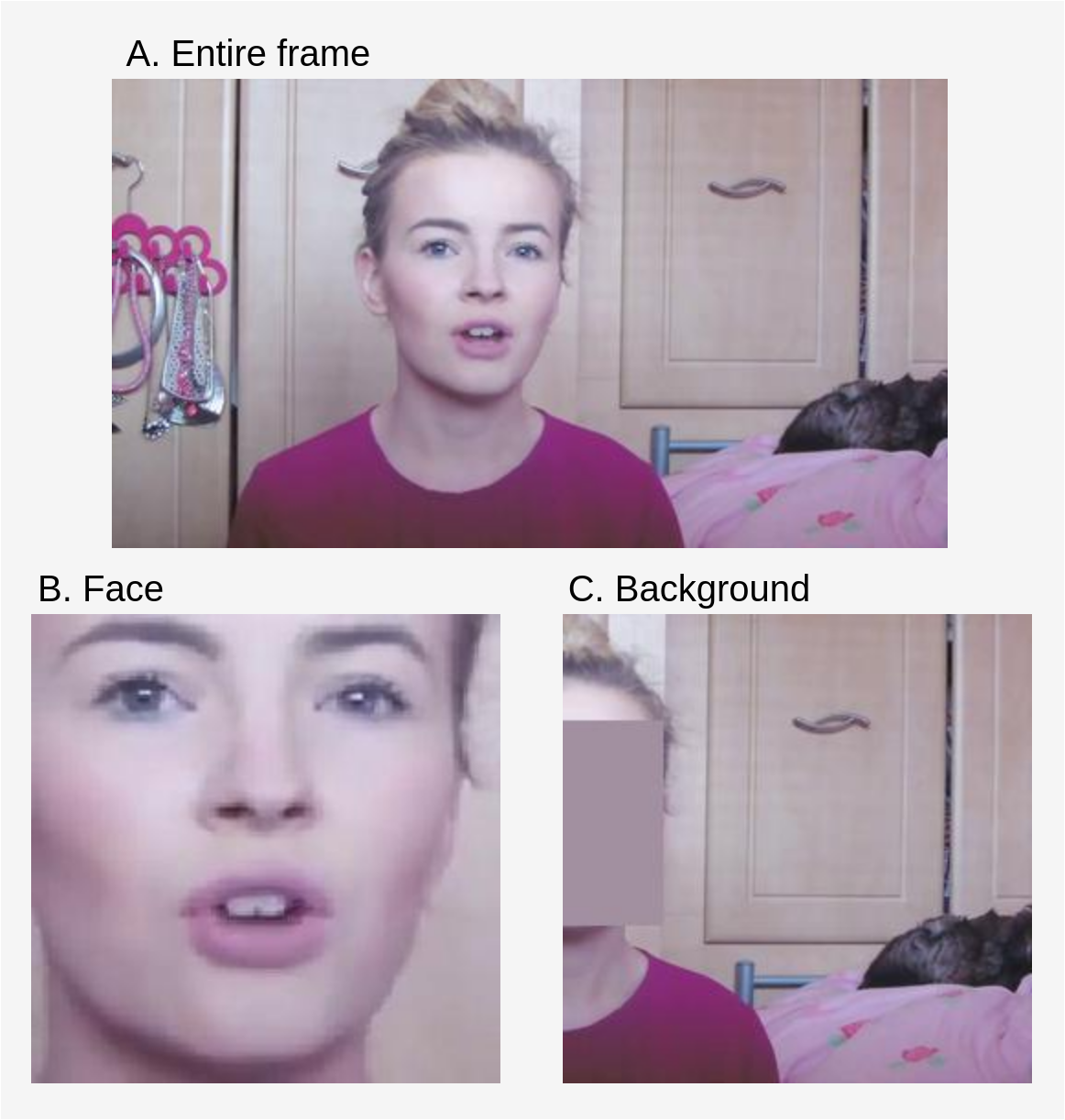}
    \caption{An overview of the different regions of data that can be given as input, A. the entire frame, B. the face and C. the background.}
    \label{fig:better_setup}
\end{figure}

In Figure \ref{fig:better_setup} an overview is given of the different experimental conditions. In total there are four conditions. In each condition a different region of the frame is given as input during the training and evaluation of the DNN. 
\\In the \textbf{face only} condition the performance of the models is measured when they are trained and evaluated on the face alone. Face extraction was performed using the dlib library~\citep{dlib09} to detect facial landmarks and return bounding box coordinates containing the location of the face for each frame of the video. The face is cropped out and resized to $256 \times 256$ pixels. The face is aligned by aligning the facial landmarks using a similarity transform to the average location of the all facial landmarks in the training split. 
\\In the \textbf{background only} condition the performance of the models is measured when they are trained and evaluated on the background alone. First the facial bounding box area is filled with the mean RGB pixel value of the image, excluding the values of the region included in the facial bounding box. To capture as much background as possible and as little person information as possible, a $256 \times 256$ crop is made starting from either the left or right edge of the image, depending on the location of the face. For example, if the face is more to the left, the crop starts from the right edge of the image. By removing the facial information completely from the frame, we can investigate if something other than the face is additionally driving the predictions. This why the body was not removed from the frame.
\\In the \textbf{face+bg} condition the performance of the models is measured when they are trained and evaluated on the face and background data.
\\Finally, the \textbf{entire frame} condition serves as a control. In this condition we measure the performance of the models when they are trained and evaluated on the entire frame. The entire frame has a size of $256 \times 465$ pixels.

\subsection{Deep Neural Networks}
Three different DNNs are used in the experiments: Deep Impression~\citep{guccluturk2016deep} and two versions of ResNet18~\citep{he2016deep}, one pre-trained on the ImageNet dataset and one without pre-training. The DNNs that are utilized in this paper belong to the widely used residual architecture family of DNNs. Three versions of the same type of architecture are used to rule out that differences in the findings are not a result of the architecture-specific components.

\subsubsection{Deep Impression}
A modified version of Deep Impression network~\citep{guccluturk2016deep} is used. The network was implemented in Chainer 4.0.0~\citep{chainer_learningsys2015}. Single frames are used to study the effect of only the background on the prediction since multiple frames bring on the effect of time on the predictions. Given that only visual information is taken into account, only the visual stream of the original network is used. The visual stream is a 17 layer deep residual network. The networks are optimized using Adam with $\alpha=0.0002$, $\beta_1 = 0.5$, $\beta_2 = 0.999$, $\epsilon = 10^{-8}$ and minibatch size of 32.

\subsubsection{ResNet18}
The ResNet18 networks are obtained from PyTorch\footnote{\url{https://pytorch.org/}}. The final layer is replaced with an fully connected layer with five outputs, one output for each B5 trait. Two ResNets are used: The first model, ResNet18 v1, is not pre-trained. 
This model, initialized with random weights, is later trained on the ChaLearn dataset. The second version, ResNet18 v2, is pre-trained on ImageNet and is fine-tuned on the ChaLearn dataset. This way we can also investigate the effects of pre-training on model performance. Pre-trained models should have a greater representational capacity and should be able to better learn relationships between background and personality traits. Both ResNet18s are optimized using stochastic gradient descent and a learning rate of 0.001 and a momentum of 0.9.

\subsubsection{Training and Validation}
Each DNN is trained on ChaLearn First Impressions v2 dataset to predict the value of the traits $OCEA\bar{N}$. $\bar{N}$ stands for $1 - N$, since the inverse of the notion of \textit{neuroticism} was used when collecting the data. All networks were trained for 100 epochs using the mean absolute error (MAE) as loss function. During the training phase, one random frame is sampled from each videoclip in the train data. After every 10 epochs the network is ran against the validation data to determine the validation loss. 
\\
\textbf{Face, background, entire frame} The networks are initialized according to the specifications given previously. 
\\
\textbf{Face+bg} The face and background images being fed to the network come from the same frame. The face stream and the background stream of the network are initialized with the weights of the \textit{face only} and \textit{background only} models. The weights are chosen from the models that perform best on the validation data. The weights of both branches are then frozen and only the final fusion layer is trained. This allows the network to update the weights accordingly to either use or ignore information coming from the branches. 

\subsubsection{Testing}
The model that has the lowest validation loss is chosen to run against the test set. Each chosen model is run against all frames in the test split and the average prediction per video per trait is recorded. Then all predictions are compared to the ground truth and the Pearson correlation coefficient is computed. In Figure \ref{fig:main_chart} the predictions per trait are averaged and compared to the mean ground truth score, also averaged across traits, and then the correlations are computed. 

\subsubsection{Comparing model performance}
Models are compared to each other in a pairwise manner, i.e. model 1 is compared to model 2. In order to determine whether the performance difference between models is meaningful the Pearson correlation coefficients $\rho$ between model predictions $X$ and the ground truth labels $Y$ are calculated:
\begin{equation}
\label{eq:pearson}
\rho = \frac{\text{cov}(X,Y)}{\sigma_x \sigma_y}
\end{equation}
These values are plotted in Figure \ref{fig:main_chart}. The Fisher transformation in \eqref{eq:z_obs} is used to obtain the $p$-values \eqref{eq:z_to_p} from the $\rho$-values. The $p$-value  measures if the difference between correlations is significant. The $p$-values are documented in Table \ref{tab:details}.
\begin{equation}
    \label{eq:z_to_p}
    p = \frac{1}{2}\Big[1+erf\Big(\frac{z_{obs}-\mu}{\sigma\sqrt{2}}\Big)\Big]
\end{equation}
where $\mu = 0$, $\sigma=1$, $erf$ is the error function and $z_{obs}$ is defined as:
\begin{equation}
    \label{eq:z_obs}
    z_{obs} = \frac{z'_1 - z'_2}{\sigma_{z'_1-z'_2}}
\end{equation}
where $z'_1=arctanh(\rho_1)$ and $z'_2=arctanh(\rho_2)$. $\sigma_{z'_1-z'_2}$ is the standard error and is defined as: 
\begin{equation}
    \label{eq:sigma_distance}
    \sigma_{z'_1-z'_2} = \sqrt{\frac{1}{N_1-3} + \frac{1}{N_2-3}}
\end{equation}
where $N_1$ is the number of pairs of scores in $\rho_1$ and $N_2$ is the number of pairs of scores in $\rho_2$. In our case $N_1=N_2=1676$ for the size of the test set. The significance level is set at $\alpha=0.05$. After Bonferroni correction $\alpha=0.05/3=0.0167$. $\alpha$ is divided by three because three different models are used; the more models we experiment with, the larger the possibility that we find a result by chance. The difference in performance is significant when $p < \alpha$. 
\\
MAE is not used directly for model comparison because the results can be misleading; a reasonable-seeming MAE (0.12) can be achieved simply by calculating the mean of the training scores. Furthermore we noticed that MAE can be very low (accuracy is very high) while correlations are very low, indicating that the model predictions are not correlated to the actual ground truth labels yet achieving a high accuracy. Further investigation shows that this is especially the case when the values of the ground truth labels lie in a very small range which is the case with the \textit{agreeableness} trait.
\section{Experimental Results}
\label{results}

\begin{figure}
    \centering
    \includegraphics[width=1.0\linewidth]{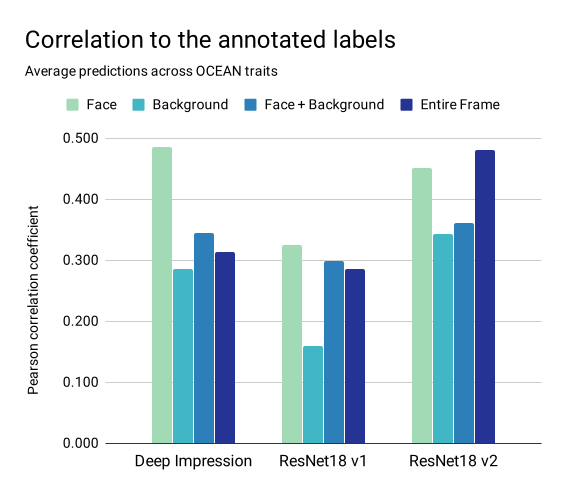}
    \caption{Mean correlation between predictions and annotated attribution labels, in each data mode, across all personality traits, for the three models that were investigated. All of the $\rho$-values computed have a $p<0.00001$, much smaller than significance level of $\alpha=0.05$ indicating that our results are likely not an effect of chance. The significance of the difference between data modes within model is documented in Table \ref{tab:details}.}
    \label{fig:main_chart}
\end{figure}

In the experiments three different DNNs are run in four experimental conditions. It is important to note that our focus lies on the performance difference between the different conditions. The absolute performance of the different models is not of interest in this paper.
We expected that the inclusion of background information, in addition to facial information, would increase the performance across all models. However this is not reflected in our results. The following observations can be made about the results presented in Figure \ref{fig:main_chart} and Table \ref{tab:details}. 
\\
\textbf{Similarity} The DNNs in the \textit{face} condition always results in a relatively high correlation with the annotated labels. This can easily be explained by the fact that faces are naturally more similar to each other compared to the set of images featuring various backgrounds, i.e. human faces have the same structure whereas backgrounds can contain various different objects that can appear in a number of different configurations. We can quantify similarity by calculating the standard deviation $\sigma$ from the mean image of the data in the different experimental conditions; the lower $\sigma$, the more similar the images are in the particular experimental condition. This results in $\sigma_{face} = 54.1$ and $\sigma_{bg} = 71.6$ for the \textit{face} and \textit{background} condition respectively. The images in the \textit{face} condition have also been aligned such that the location of the facial landmarks is the same for all the images. This makes the learning task easier for the DNNs. In contrast nothing has been done to increase the "structuredness" of the images in the \textit{background} condition.
\\
\textbf{SNR} When background information is explicitly added to the input, i.e. \textit{face} vs. \textit{face+bg} condition, a relative {\em decrease} in correlation was measured across all models in comparison to the \textit{face} condition. This decrease in correlation is significant for the Deep Impression and ResNet18 v2 models. Initially this may seem like a surprising result because the DNNs in the \textit{background} condition do show some correlation to the ground truth. In theory it means that when more data is given to the model, the better it should perform, especially since both conditions show correlation to the ground truth. However, the correlation in the \textit{background} condition is significantly lower compared to the correlation in the \textit{face} condition, see Table \ref{tab:details}. This performance gap indicates that, for the tested DNN architectures, the \textit{face} condition is significantly more informative than the \textit{background} condition. From an information theoretical perspective the analogy of the signal-to-noise ratio (SNR) can be made; we can say that the \textit{background} condition contains more noise, decreasing the correlation when added to the \textit{face} condition, $\rho_{face+bg} < \rho_{face}$. And inversely that the \textit{face} condition contains more signal, increasing the correlation when added to the \textit{background} condition, $\rho_{face+bg} > \rho_{background}$. 
\\
\textbf{Pre-train} It can be observed that the correlations of ResNet18 v2 are relatively high for all conditions. Given that the model has been pre-trained on ImageNet, we can assume that this pre-training is increasing the model's ability to find useful features, both in the background and the face. Even though this DNN behaves similar to the other two DNNs in the \textit{face+bg} condition, it performs much better in the \textit{entire frame} condition. We suspect that the combination of pre-training and the fact that the entire frame contains more information are causing the increase in correlation. This suspicion is also reflected in the fact that the other two DNNs have not been pre-trained on ImageNet and that their performance in the \textit{entire frame} condition is relatively low compared to the \textit{face} condition.

\begin{table}[!h]
\renewcommand{\arraystretch}{1.3}
\caption{An overview of the comparisons between correlations of Figure \ref{fig:main_chart}. The data modes that differ significantly have a $p<\alpha$ and are indicated with an asterisk. $\alpha=0.0167$. This table should be viewed in conjunction with Figure \ref{fig:main_chart}.}
\centering
\begin{tabular}{|l|c|c|}

\hline
\textbf{Deep Impression} & \textbf{$z_{obs}$} & \textbf{$p$} \\ \hline
face vs. face+bg $*$        & $4.91$ & $4.45e-7$ \\
face vs. entire frame $*$   & $5.96$ & $1.26e-9$ \\ 
face vs. bg $*$             & $6.83$ & $4.06e-12$ \\
bg vs. face+bg              & $1.93$ & $0.027$ \\ 
bg vs. entire frame         & $0.89$ & $0.19$ \\
face+bg vs. entire frame    & $1.04$ & $0.148$ \\\hline

\textbf{ResNet18 v1} & \textbf{$z_{obs}$} & \textbf{$p$} \\ \hline
face vs. face+bg            & $0.85$ & $0.198$ \\
face vs. entire frame       & $1.27$ & $0.103$ \\ 
face vs. bg $*$             & $5.08$ & $1.76e-7$\\ 
bg vs. face+bg $*$          & $4.22$ & $1.10e-5$ \\ 
bg vs. entire frame $*$     & $3.81$ & $6.47e-5$ \\
face+bg vs. entire frame    & $0.41$ & $0.339$ \\ \hline

\textbf{ResNet18 v2} & \textbf{$z_{obs}$} & \textbf{$p$} \\ \hline
face vs. face+bg $*$        & $3.11$ & $9.35e-4$ \\
face vs. entire frame       & $-1.1$ & $0.136$ \\
face vs. bg $*$             & $3.71$ & $9.51e-5$ \\
bg vs. face+bg              & $0.63$ & $0.266$ \\
bg vs. entire frame $*$     & $4.82$ & $6.83e-7$ \\
face+bg vs. entire frame $*$& $-4.2$ & $1.30e-5$ \\ \hline
\end{tabular}
\label{tab:details}
\end{table}

\section{Discussion}
Technologies that make use of facial information remain a controversial topic in general. However, researching said technology and the methods behind them is beneficial when the goal is to understand the inner workings and assess the reliability of these methods given that they are being used in the real world to make actionable decisions.

In this study, we considered the visual sources of information that drive apparent personality attribution using DNNs. Three different DNNs were run in four conditions of the ChaLearn First Impressions v2 dataset. It was expected that the inclusion of background information, in addition to facial information, would increase the performance across all models. 

Surprisingly, we found no evidence that background information is improving model attributions for apparent personality traits. In fact, when background is explicitly added to the input, a decrease in performance was measured across all models. Our results do suggest that correlations to ground truth can be boosted by training on the entire frame, but the result is not significantly higher than training on the facial information alone. 
From the experiments we can conclude that facial information is significantly more informative to our models than the background information, even when the model is pre-trained on ImageNet and given access to the complete frame information. However, it is notable that, in the \textit{background} condition, the DNNs can perform trait attributions with some correlation to the labels. This suggests that there is a regularity present in the \textit{background} condition that the DNNs pick up on. Further investigation is required to determine to what degree the human annotators could utilize information in the background and if they have done so.

As is often the case with deep learning applications, it is difficult to say with any certainty whether the results will replicate across other architectures and datasets. DNN architectures are by their very nature incredibly complex structures containing many interactive components. The net result of the interaction between these components cannot be predicted in advance~\citep{ras2018explanation}. To address this limitation the experiments should be performed on more DNN architectures. The ability of the model to extract informative features from the data is crucial to its performance as has been shown by pre-training the network. The recent advances in contextual feature extraction indicate that it is possible to create networks with a human-like capability of image understanding for certain tasks, especially image captioning~\citep{hossain2019comprehensive}. However, we still have a long way to go to bridge the gap to subjective social human understanding from image data. 

Concluding, our results suggest that DNNs mainly exploit facial features to predict apparent personality traits. Future research should provide further insights into how exactly facial features determine particular apparent personality traits.

\bibliographystyle{apalike}
\bibliography{sample}

\end{document}